\documentclass[runningheads,a4paper]{llncs}

\usepackage{amssymb}
\usepackage{graphicx}
\usepackage{url}
\urldef{\mailsa}\path|{rajesh.dm, gsr}@iiitb.ac.in|    
\newcommand{\keywords}[1]{\par\addvspace\baselineskip
\noindent\keywordname\enspace\ignorespaces#1}
\usepackage{hyperref}
\usepackage{url}
\usepackage{algorithm}
\usepackage{algpseudocode}
\usepackage{graphicx}
\usepackage[export]{adjustbox}

\begin{document}

\mainmatter  % start of an individual contribution

\title{Learning  Agents With Prioritization and Parameter Noise in Continuous State and Action Space}

\titlerunning{Prioritization and Parameter Noise for Better RL Agents}%%%%%%%%%%%

%\author{Alfred Hofmann%
%\thanks{Please note that the LNCS Editorial assumes that all authors have used
%the western naming convention, with given names preceding surnames. This determines
%the structure of the names in the running heads and the author index.}%
%\and Ursula Barth\and Ingrid Haas\and Frank Holzwarth\and\\
%Anna Kramer\and Leonie Kunz\and Christine Rei\ss\and\\
%Nicole Sator\and Erika Siebert-Cole\and Peter Stra\ss er}

\author{Rajesh Mangannavar and Gopalakrishnan Srinivasaraghavan}

\authorrunning{Prioritization and Parameter Noise for Better RL Agents}

\institute{International Institute of Information Technology, Bangalore\\
Bangalore, India \\
\mailsa}


%\toctitle{Lecture Notes in Computer Science}
%\tocauthor{Authors' Instructions}
\maketitle

\begin{abstract}
 Among the many variants of RL, an important class of problems is where the state and action spaces are continuous --- autonomous robots, autonomous vehicles, optimal control are all examples of such problems that can lend themselves naturally to reinforcement based algorithms, and have continuous state and action spaces. In this paper, we introduce a prioritized form of a combination of state-of-the-art approaches such as Deep Q-learning (DQN) and Deep Deterministic Policy Gradient (DDPG) to outperform the earlier results for continuous state and action space problems. Our experiments also involve the use of parameter noise during training resulting in more robust deep RL models outperforming the earlier results significantly. We believe these results are a valuable addition for continuous state and action space problems.

\end{abstract}

\section{Introduction}

Reinforcement learning (RL) is about an agent learning an optimal way to control and/or navigate through an environment that requires sequential decision making by the agent. The agent does this by trying to maximize a numerical performance measure that expresses a long-term objective, by trial-and-error. RL arises naturally in a wide range of domains including autonomous control, gaming, natural language processing \& dialogue management, etc. \cite{DeepRLOverview_1:journals/corr/Li17b}.

\subsection{RL in Continuous State and Action Space}

Many interesting real-world control tasks, such as driving a car or riding a snowboard, require smooth continuous actions taken in response to high-dimensional, real-valued sensory inputs. 
In applying RL to continuous problems, the predominant approach in the past involved 
discretizing the state and action spaces and then applying an RL algorithm for a discrete stochastic system \cite{RL_in_cont_time}. However the drawbacks of discretization are, they do not scale, do not allow fine-grained smooth control characteristic of continuous space-action systems, and are too sensitive to the (arbitrary) choice of the granularity with which the discretization is carried out.

Hence, the formulation of the reinforcement learning problem with continuous state and action space holds great value in solving more real-world problems.

\subsection{Deep Reinforcement Learning}
The advent of deep learning has had a significant impact
in many areas in machine learning, dramatically improving
the state-of-the-art tasks such as object detection, speech
recognition and language translation \cite{DeepLearning}. Deep Neural Networks are very powerful function approximators and can be trained to automatically find low-dimensional representations of high-dimensional data. This enables deep learning to scale to problems which were previously intractable including reinforcement learning problems with high dimensional, continuous state and action spaces \cite{DeepRLOverview_2:journals/corr/abs-1708-05866}. 

Few of the current state of the art methods in the area of Deep RL are: 

\begin{itemize}
    \item Deep Q-learning Networks (DQN) - introduced novel concepts which helped in using neural networks as function approximators for reinforcement learning algorithms (for continuous state space)    \cite{DQN:journals/corr/MnihKSGAWR13}.
    
    \item Prioritized Experience Replay (PER) - builds upon DQN with some newer approaches to outperform DQN  (for continuous state space)  \cite{PAR:journals/corr/SchaulQAS15}.
    
    \item Deep Deterministic Policy Gradients (DDPG) - follows a different paradigm as compared to the above methods. It uses DQN as the function approximator while building on the seminal work of \cite{DPG:Silver:2014:DPG:3044805.3044850} (for both continuous state and action space) on deterministic policy gradients.  
\end{itemize}

\subsection{Prioritized Experience Replay in DDPG ( PDDPG)}

We propose a new algorithm, Prioritized DDPG using the ideas proposed in DQN, prioritized experience replay and DDPG such that it outperforms the DDPG in the continuous state and action space. Prioritized DDPG uses the concept of prioritized sampling in the function approximator of DDPG. 
Our results show that prioritized DDPG outperforms DDPG in a majority of the continuous action space environments. We then use the concept of parameter space noise for exploration and show that this further improves the rewards achieved.

\section{Previous work}
\label{prev_work}

\subsection{Critic Methods of RL}\label{sec:critic}

Critic methods rely exclusively on a value or Q-function approximation and aim at learning a ``good'' approximation of the value/Q-function\cite{actor_critic_algorithms}. We survey a few of the recent best-known critic methods in RL.

\subsubsection{Deep Q-learning Networks}\label{sec:DQN}

As in any value function-based approach, DQN method tries to find the optimal value function for any given state / state-action-pair. The novelty of their approach is that they efficiently use a non-linear function approximator to learn the function. Prior to their work, though the potential for using non-linear function approximators was recognized, there was no tangible demonstration of their use in practice in an efficient manner.

The challenge in using function approximators was that the Monte-Carlo sampling typically used for collecting experience in the form of state-action-rewards by generating episodes did not guarantee i.i.d data that most machine learning algorithms assume for training the models from.  To overcome this drawback, the novel ideas that were  proposed which made it possible to use non-linear function approximators for reinforcement learning are the following:

\begin{itemize}
    \item Experience Replay Buffer: In this technique, the neural network is trained from random samples from a large buffer of stored observations. 
    
    \item Periodic Target Network Updates: Two sets of parameters are maintained for the same neural network --- one for generating behaviour from (possibly using an $\epsilon$-greedy strategy) and the other (the target network) for the value function estimation. The target network parameters are used in the loss computation and are updated by the behaviour network parameters periodically.
    
\end{itemize}

\subsubsection{Prioritized Experience Replay}\label{sec:PAR}

The prioritized experience replay algorithm is a further improvement on the deep Q-learning methods and can be applied to both DQN and Double DQN. The idea proposed by the authors is as follows: instead of selecting the observations at random from the replay buffer, they can be chosen based on some criteria which will help in making the learning faster. Intuitively the selection criterion from the replay buffer is biased towards the 'more useful observations' and less on the 'stale' ones. To select these more useful observations, the criteria they use is the error of that particular observation. 

This criterion helps select those observations which provide the highest learning opportunity for the agent. The problem with this approach is that greedy prioritization focuses on a small subset of the experience and this lack of diversity may lead to over-fitting. Hence, the authors introduce a stochastic sampling method that interpolates between pure greedy and uniform random prioritization. Hence, the new probability of sampling a transition $i$ is 

\begin{equation}\label{eqn:stochastic_sampling}
P(j) = \frac{p^{\alpha}_j}{\sum\nolimits_{k}p^{\alpha}_k}
\end{equation}

where $p_i$ is the priority of transition $i$ and $\alpha$ determines how much prioritization is used. The approach, while it improves the results has a problem of changing the distribution of the expectation. This is resolved by the authors by using Importance Sampling (IS) weights 

\begin{equation}\label{eqn:importane_sampling}
w_i = \bigg(\frac{1}{N}\cdot\frac{1}{P(i)}\bigg)^\beta
\end{equation}

where if $\beta=1$, the non-uniform probabilities $P(i)$ are fully compensated (\cite{PAR:journals/corr/SchaulQAS15}).

\subsection{Actor Methods in RL}\label{sec:actor}

Actor methods work with a parameterized family of policies. The gradient of the performance, with respect to the actor parameters, is directly estimated by simulation, and the parameters are updated in the direction of improvement (\cite{actor_critic_algorithms}).
%define actor methods

\subsubsection{Deterministic Policy Gradients (DPG)}

The most popular policy gradient method is the deterministic policy gradient(DPG) method and in this approach, instead of having a stochastic policy, the authors make the policy deterministic and then determine the policy gradient.

The deterministic policy gradient is the expected gradient of the
action-value function, which integrates over the state space, whereas in the stochastic case, the policy gradient integrates over both state and action spaces. What this leads to is that the deterministic policy gradient can be estimated more efficiently than the stochastic policy gradient.

The DPG algorithm, presented by \cite{DPG:Silver:2014:DPG:3044805.3044850} maintains a parameterized actor function $\mu (s|\theta^{\mu})$ which is the current deterministic policy that maps a state to a particular action. The authors use the Bellman equation to update the critic $Q(s, a)$. They then go on to prove that the derivative expected return with respect to actor parameters is the gradient of the policy's performance.

\subsection{Actor-Critic Methods}\label{sec:actor_critic}

Actor-critic models (ACM) are a class of RL models that separate the policy from the value approximation process by parameterizing the policy separately. The parameterization of the value function is called the critic and the parameterization of the policy is called the actor. The actor is updated based
on the critic which can be done in many ways, while the critic is update based on the current policy provided by the actor (\cite{actor_critic_algorithms} \cite{Handbook_of_markov}).

\subsubsection{Deep Deterministic Policy Gradients (DDPG)}\label{sec:DDPG}

The DDPG algorithm tries to solve the reinforcement learning problem in continuous action and state space setting. The authors of this approach extend the idea of deterministic policy gradients. What they add to the DPG approach is the use of a non-linear function approximator (\cite{DDPG:journals/corr/LillicrapHPHETS15}). 

While using a deterministic policy, the action value function reduces from 

\begin{equation}
Q^{\pi}(s_t,a_t) = \mathbb{E}_{r_{t},s_{t+1} \sim E}[r(s_t, a_t) + \gamma\mathbb{E}_{a_{t+1} \sim \pi }[Q^{\pi} (s_{t+1}, a_{t+1} )]]
\end{equation}

\noindent
to 

\begin{equation}
Q^{\mu}(s_t,a_t) = \mathbb{E}_{r_{t},s_{t+1} \sim E}[r(s_t, a_t) + \gamma Q^{\mu} (s_{t+1}, \mu( s_{t+1}) )]
\end{equation}

\noindent
as the inner expectation is no longer required. Here, $\gamma \in [0,1]$ is the discounting factor.  What this also tells us is that the expectation depends only on the environment and nothing else. Hence, we can learn off-policy, that is, we can train our reinforcement learning agent by using the observations made by some other agent. The authors use this as well as the novel concepts used in DQN to construct their function approximator. 
These concepts cannot be applied directly to continuous action space, as there is an optimization over the action space at every step which is in-feasible when there is a continuous action space \cite{DDPG:journals/corr/LillicrapHPHETS15}.

Once we have both the actor and the critic networks with their respective gradients, we can then use the DQN concepts - replay buffer and target networks to train these two networks. They apply the replay buffer directly without any modifications but make small changes in the way target networks are used. Instead of directly copying the values from the temporary network to the target network, they use soft updates to the target networks. 

\subsubsection{Parameter Space Noise for Exploration}

There is no best exploration strategy in RL. For some algorithms, random exploration works better and for some greedy exploration. But whichever strategy is used, the important requirement is that the agent has explored enough about the environment and learns the best policy. Plappert et. al.  \cite{parameter_noise:journals/corr/PlappertHDSCCAA17} in their paper explore the idea of adding noise to the agent's parameters instead of adding noise in the action space. In their paper Parameter Space Noise For Exploration, they explore and compare the effects of four different kinds of noises \begin{itemize}
    \item Uncorrelated additive action space noise
    \item Correlated additive Gaussian action space noise
    \item Adaptive-param space noise
    \item No noise
\end{itemize}

They show that adding parameter noise vastly outperforms existing algorithms or at least is just as good on a majority of the environments for DDPG as well as other popular algorithms such as Trust Region Policy Optimization (\cite{TRPO:journals/corr/SchulmanLMJA15}). 

\section{Prioritized Deep Deterministic Policy Gradients}

The proposed algorithm is primarily an adaptation of DQN and DDPG with ideas from the work of Schaul et. al. \cite{PAR:journals/corr/SchaulQAS15} on continuous control with deep reinforcement learning to design an RL scheme that improves on DDPG significantly.
The intuition behind the idea is as follows: The DDPG algorithm uses the DQN method as a sub-algorithm and any improvement over the DQN algorithm should ideally result in the improvement of the DDPG algorithm.
But from the above-described methods, not all algorithms which improve DQN can be used to improve DDPG. That is because some of them need the environment to have discrete action spaces. So, for our work, we will consider only the prioritized experience replay method which does not have this constraint. 

\subsection{ Prioritized DDPG Algorithm }

Now, the improvement to the DQN algorithm, the prioritized action replay method can be integrated into the DDPG algorithm in a very simple way. Instead of using just DQN as the function approximator, we can use DQN with prioritized action replay. That is, in the DDPG algorithm, instead of selecting observations randomly, we select the observations using the stochastic sampling method as defined in equation \ref{eqn:stochastic_sampling}. 
The pseudo-code for the prioritized action replay is given in Algorithm
\ref{PDDPGalgorithm}. The algorithm runs in a for loop $M$ times where $M$ is the number of episodes we want to train the agent for.

\begin{algorithm}[!htb]
    {\fontfamily{pcr}\selectfont 
    \caption{PDDPG algorithm}}
    \label{PDDPGalgorithm}
    \begin{algorithmic}
    
    {\fontfamily{pcr}\selectfont 
        \State Randomly initialize critic network $Q(s,a|\theta^Q)$ and actor $\mu(s|\theta^\mu)$ with weights $\theta^Q$ and $\theta^\mu$.
        \State Initialize target network $Q'$ and $\mu'$ with weights $\theta^{Q'} \leftarrow \theta^{Q}$,$\theta^{\mu'} \leftarrow \theta^{\mu}$
        \State Initialize replay buffer $R$
    
        \For{ episode = 1, M}
            \State Initialize a random process \textit{N} for action exploration
            \State Receive initial observation state $s_1$
            \For{t = 1, T}
                \State Select action $a_t = \mu (s_t|\theta^\mu) + \textit{N}_t$ using to the current policy
                \State Execute action $a_t$ and observe reward $r_t$ and new state $s_{t+1}$
                \State Store transition $(s_t , a_t , r_t , s_{t+1} )$ in $R$  \Comment{Storing to the replay buffer}
                \State{Sample a mini-batch of \textit{N} transitions $(s_i , a_i , r_i , s_{i+1} )$ from $R$ each such that - $P(i) = p_i^{\alpha}/\Sigma_i p_i^{\alpha}$}  \Comment{Stochastic sampling}
                \State Set $y_i = r_i + \gamma Q^{'}(s_{i+1}, \mu^{'}(s_{i+1}|\theta^{\mu^{'}})|\theta^{Q^{'}})$
                \State Update critic by minimizing the loss: $L = \frac{1}{N} \sum_i(y_{i} - Q(s_i, a_i|\theta^{Q}))^2$
                \State Update the actor policy using the sampled policy gradient
                \begin{center}
                    \State $ \nabla_{\theta^\mu}J \approx \frac{1}{N} \sum_i\nabla_aQ(s,a|\theta^{Q})|_{s=s_i,a=\mu(s_i)}\nabla_{\theta^\mu}\mu(s|\theta^{\mu})|_{s_i}$
                    \State
                \end{center}
                \State Update the target networks:
                \begin{center}
                    \State  $\theta^{Q^{'}} \leftarrow \tau \theta^Q+(1-\tau)\theta^{Q^{'}}$
                    \State  $\theta^{\mu^{'}} \leftarrow \tau \theta^\mu+(1-\tau)\theta^{\mu^{'}}$
                \end{center}
                \State Update the transition priorities for the entire batch based on the error  
            \EndFor
        \EndFor
        }
    \end{algorithmic}
    \end{algorithm}

This algorithm is quite similar to the original DDPG algorithm with the only changes being the way the observations are selected for training and the transition probabilities are being updated. The first change ensures we are selecting the better set of observations which help in learning faster and the second change helps in avoiding over-fitting as it ensures all the observations have a non-zero probability of being selected to train the network and only a few high error observations are not used multiple times to train the network. 

\section{ PDDPG With Parameter Noise }

As introducing parameter noise improved the results obtained by DDPG, we introduce the parameter noise with PDDPG in the same way. The noise is added such we can achieve structured exploration by applying the noises to the parameter of the current policy. Also, the policy on which we have applied our noise is sampled at the beginning of each episode. 

The noise we add to the parameter are the ones discussed before - 

\begin{itemize}
    \item Uncorrelated additive action space noise
    \item Correlated additive Gaussian action space noise
    \item Adaptive-param space noise
\end{itemize}

We also have a result without adding any noise (original PDDPG) for comparison. 

\section{Results}

The proposed, prioritized DDPG algorithm was tested on many of the standard RL simulation environments that have been used in the past for benchmarking the earlier algorithms. The environments are available as part of the Mujoco platform (\cite{Mujoco}).

\subsection{Mujoco Platform}\label{sec:mujoco}

Mujoco is a physics environment which was created to facilitate research and development in robotics and similar areas, where fast simulation is an important component.

This set of environments provide a varied set of challenges for the agent as environments have continuous action as well as state space. All the environments contain stick figures with some joints trying to perform some basic task by performing actions like moving a joint in a particular direction or applying some force using one of the joints.

\subsection{Empirical Evaluation}

The implementation used for making the comparison was the implementation of DDPG in baselines (\cite{baselines}). The prioritized DDPG algorithm was implemented by extending the existing code in baselines.

\subsubsection{Results for PDDPG }
The following are the results of the prioritized DDPG agent as compared DDPG agent (\cite{DDPG:journals/corr/LillicrapHPHETS15}). The overall reward - that is the average of the reward across all epochs until that point and reward history - the average of the last 100 epochs on four environments are plotted. 
The y-axis represents the reward the agent has received from the environment and the x-axis is the number of epochs with each epoch corresponding to $2000$ time steps.

\begin{figure}[!htb]
\begin{center}
\includegraphics[width=\linewidth]{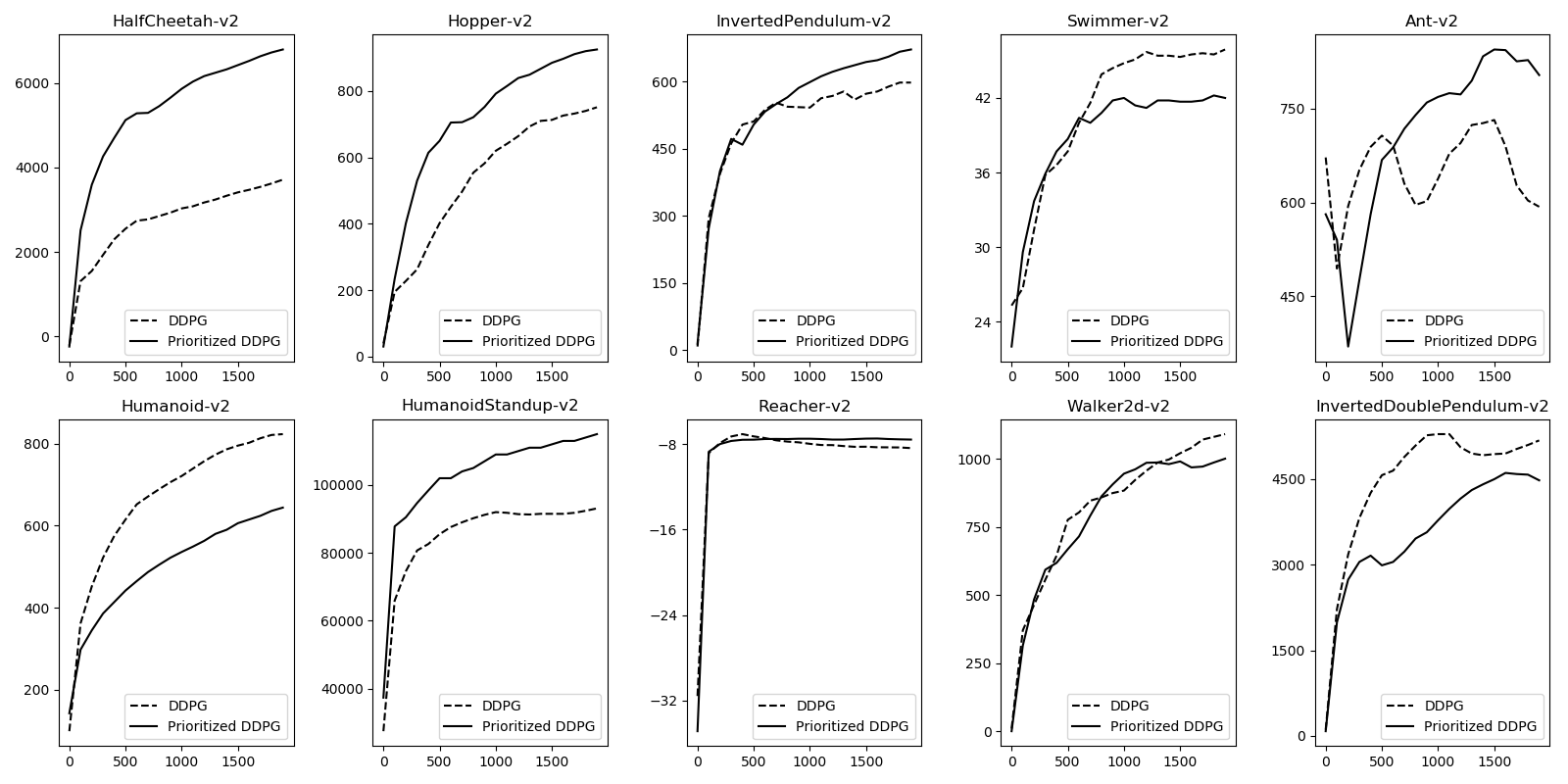}
\end{center}
\caption{Prioritized DDPG vs DDPG}
\label{no_noise}
\end{figure}

As seen in Figure \ref{no_noise}, the Prioritized DDPG algorithm reaches the reward of the DDPG algorithm in less than 300 epochs for the HalfCheetah environment. This shows that the prioritized DDPG algorithm is much faster in learning.

The same trend can be observed in Figure \ref{no_noise} for HumanoidStandup, Hopper and Ant environments. That is,the prioritized DDPG agent learns and gets the same reward as DDPG much faster. This helps is in reducing overall training time. Prioritized DDPG algorithm can also help in achieving results which might not be achieved by DDPG even after a    large number of epochs. This can be seen in the case of the Ant environment. Figure \ref{no_noise} shows that DDPG rewards are actually declining with more training. On the other hand, Prioritized DDPG has already achieved a reward much higher and is more stable.

One a few environments such as the Reacher, InvertedDoublePendulum and Walker2d, it can be seen from Figure \ref{no_noise} the prioritized DDPG only, marginally outperforms the DDPG algorithm.

\subsubsection{Results for PDDPG with parameter noise }

The PDDPG algorithm with parameter noise was run on the same set of environments as the PDDPG algorithm - the Mujoco environments. The empirical results are as follows.

\begin{figure}[!htb]
\begin{center}
\includegraphics[width=\linewidth]{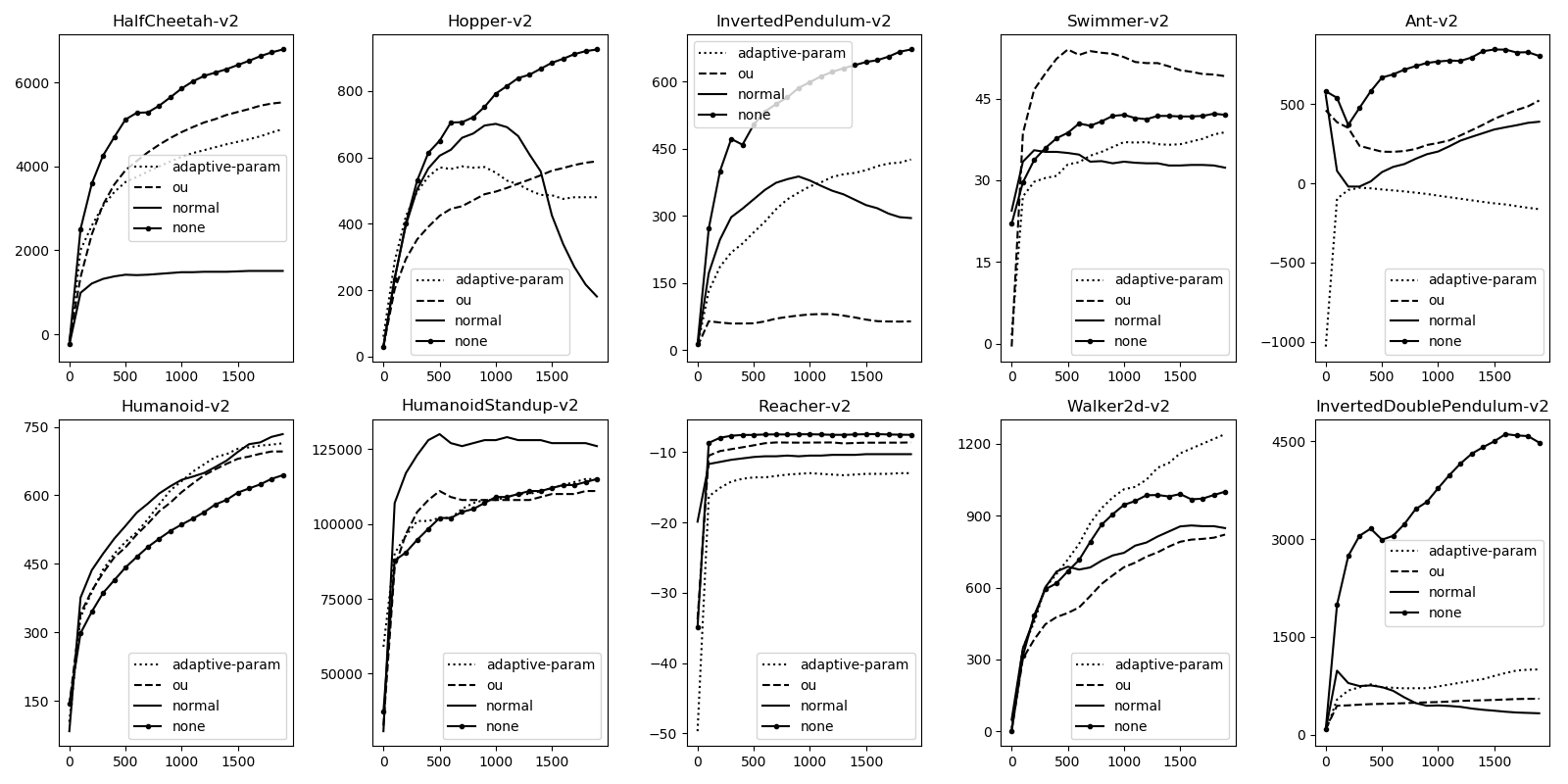}
\end{center}
\caption{Prioritized DDPG across all noi ses - adaptive-param, uncorrelated, co related and  with no noise}
\label{fig:PAR_all}
\end{figure}

We can see from figure \ref{fig:PAR_all} that noise improves the overall reward obtained only some of the environment. This is because the idea behind adding noise is for better exploration and PDDPG explores and learns much faster as seen in \ref{no_noise}. In environments such as Inverted Pendulum or inverted double pendulum where lesser exploration is required, the addition of noise does not improve the rewards achieved, but in environments such as Humanoid-Stand up or Walker-2d, where a lot of exploration is required, the noise does improve the overall reward achieved. 

\begin{figure}[!htb]
\begin{center}
\includegraphics[width=\linewidth]{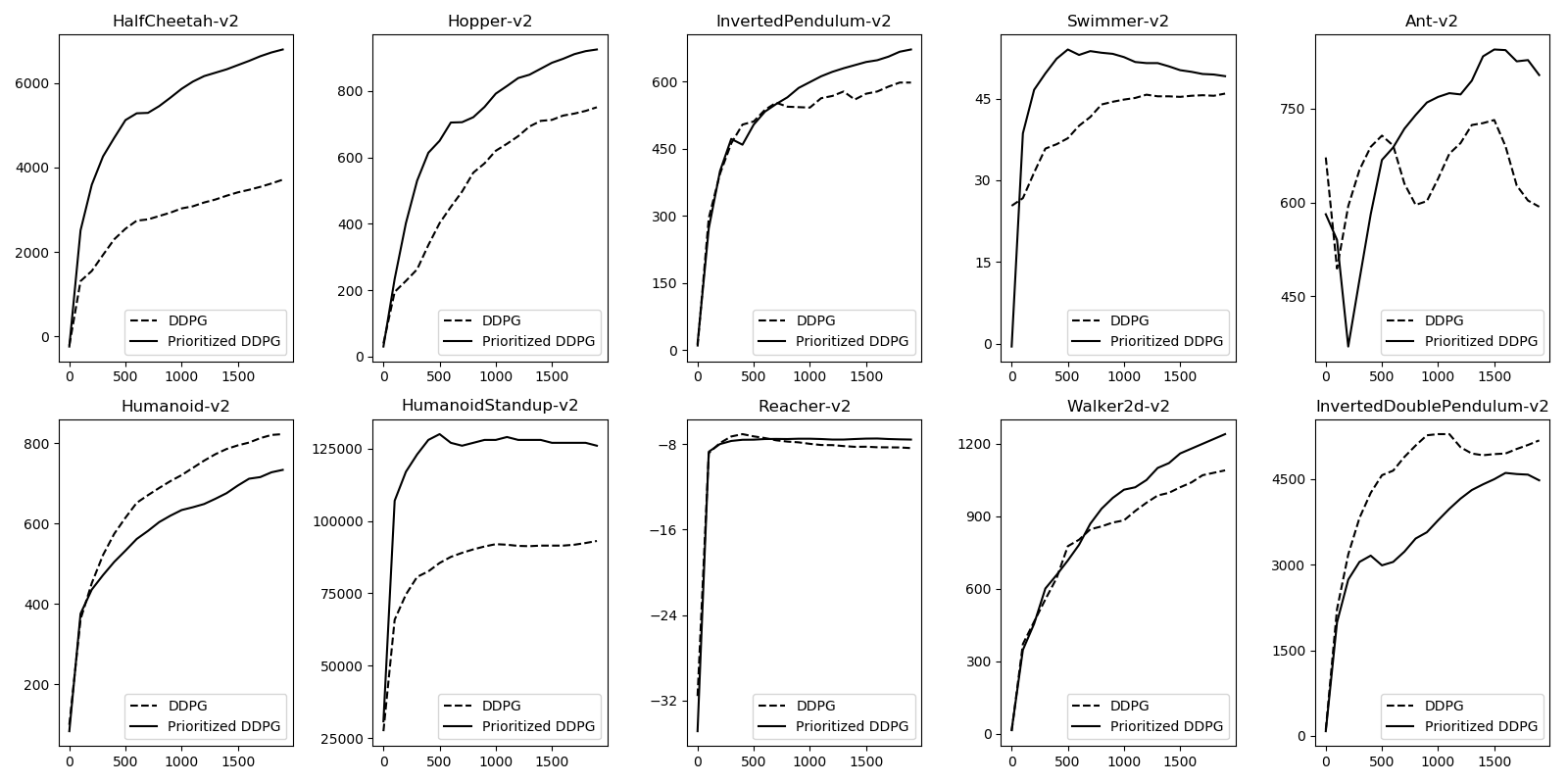}
\end{center}
\caption{Prioritized DDPG with noise vs DDPG}
\label{fig:best_of_all}
\end{figure}

Overall, with the proposed changes, prioritization with the addition of noise, the proposed algorithm outperforms DDPG on eight of the ten environments as seen in Figure \ref{fig:best_of_all} and does reasonably well in the others. 

\section{Conclusions}

To summarize, this paper discusses the state of the art methods in reinforcement learning with our improvements that have led to RL algorithms in continuous state and action spaces that outperform the existing ones.

 The proposed algorithm combines the concept of prioritized action replay with deep deterministic policy gradients. As it has been shown, on a majority of the mujoco environments this algorithm vastly outperforms the DDPG algorithm both in terms of overall reward achieved and the average reward for any hundred epochs over the thousand epochs over which both were run. 
 
 The proposed algorithm seems to learn much faster than the DDPG algorithm. Also after 2000 iterations, as the graphs above show, the proposed algorithm had accumulated significantly more rewards than DDPG while still retaining a positive slope compared to the flattened curve for DDPG. This indicates that it is unlikely for the DDPG algorithm to surpass the results of the proposed algorithm on a majority of the environments. Also, certain kinds of noises further improve PDDPG to help attain higher rewards. One other important conclusion is that different kinds of noises work better for different environments which is evident in how drastically the results changed based on the parameter noise.
 
The presented algorithm can also be extended and improved further by finding more concepts in value based methods, which can be used in policy-based methods. The overall improvements in the area of continuous space and action state space can help in making reinforcement learning more applicable in real-world scenarios.  These methods can potentially be extended to safety-critical systems, by incorporating the notion of safety during the training of an RL algorithm. This is currently a big challenge because of the necessary unrestricted exploration process of a typical RL algorithm.

%\bibliographystyle{unsrt} 
%\bibliography{bibliography}

\end{document}